\UseRawInputEncoding
\documentclass[letterpaper, 10 pt, conference]{ieeeconf}  % Comment this line out if you need a4paper

\IEEEoverridecommandlockouts                              % This command is only needed if 
\usepackage{float}
\usepackage{bm}
\usepackage{amsfonts,amssymb}
\usepackage{multirow}
\usepackage{subcaption}
\usepackage{graphicx}
\usepackage{graphics} % for pdf, bitmapped graphics files
\usepackage{url}
\usepackage{booktabs}
\usepackage{cite}
\usepackage[ruled,vlined,linesnumbered]{algorithm2e}
\usepackage{amsmath} % assumes amsmath package installed

\title{\LARGE \bf
Palm-sized Omnidirectional Vision-Based UAV Exploration\\ with Sparse Topological Map Guidance
}

\author{ Zirui Wang$^{1,2,3,*}$, Xinjia Luo$^{1,*}$, Haotian Sun$^{1}$, Jun Ma$^{2}$, Jian Guo$^{3}$, and Boyu Zhou$^{1,\dagger}$
\thanks{$^*$ Equal contribution. $^\dagger$ Corresponding Author.}
\thanks{$^1$ Department of Mechanical and Energy Engineering, Southern University of Science and Technology, Shenzhen, 518055, China 
}
\thanks{$^2$ Thrust of Robotics and Autonomous Systems, Hong Kong University of Science and Technology (Guangzhou), Guangzhou, 511453, China}
\thanks{$^3$ International Digital Economy Academy, Shenzhen, China}
}
\begin{document}

\maketitle
\thispagestyle{empty}
\pagestyle{empty}

%%%%%%%%%%%%%%%%%%%%%%%%%%%%%%%%%%%%%%%%%%%%%%%%%%%%%%%%%%%%%%%%%%%%%%%%%%%%%%%%
\begin{abstract}

Classic exploration methods often rely on dense occupancy maps or high-resolution point clouds for frontier detection and path planning, resulting in substantial memory consumption and computational overhead. Moreover, micro UAVs under size, weight, and power (SWaP) constraints are not practical to be equipped with sensors like LiDAR to obtain accurate environmental geometric measurements.
This paper presents a lightweight autonomous exploration system that leverages omnidirectional vision and sparse topological map guidance. Specifically, we utilize a multi-fisheye camera setup to achieve omnidirectional Field of View (FoV) and perform depth estimation.
To address the limited depth estimation accuracy, frontiers are represented as potential unexplored regions characterized by topological nodes instead of explicit boundaries, enabling efficient identification of frontier regions without maintaining occupancy grids or global point clouds.
Unlike classic dense representations, our approach abstracts the environment using a sparse topological map composed of key nodes and their descriptors, reducing memory consumption and computational demands. Global path planning is performed directly on the sparse graph. 
The proposed method is validated in both simulation and on a palm-sized vision-based UAV with an 11 cm wheelbase and a 400 g weight in real-world experiments, demonstrating that our method can achieve efficient exploration with extremely low computational consumption.

\end{abstract}

% \begin{keywords}
% Visual SLAM.
% \end{keywords}
%%%%%%%%%%%%%%%%%%%%%%%%%%%%%%%%%%%%%%%%%%%%%%%%%%%%%%%%%%%%%%%%%%%%%%%%%%%%%%%%

% \input{sections/introduction.tex}
% \input{sections/related_work.tex}
% \input{sections/problem_formulation.tex}
% \input{sections/system_overview.tex}
% \input{sections/algorithm.tex}
% \input{sections/results.tex}
% \input{sections/conclusion.tex}

\section{Introduction}

Autonomous exploration for unmanned aerial vehicles (UAVs) is a fundamental challenge in robotics, enabling full coverage of complex unknown environments using only onboard sensing and computing resources. While most existing methods are designed for conventional-sized UAV platforms with sufficient payload and computational power, efficient autonomous exploration for palm-sized micro UAVs under strict size, weight, and power (SWaP) constraints remains a critical  bottleneck. Importantly, such compact robots offer irreplaceable advantages in confined scenarios, including narrow-space inspection and post-disaster search and rescue, where full-size UAVs cannot gain access, making the advancement of their exploration capabilities highly valuable.

\begin{figure*}[t]
    \centering
    \includegraphics[width=.98\linewidth]{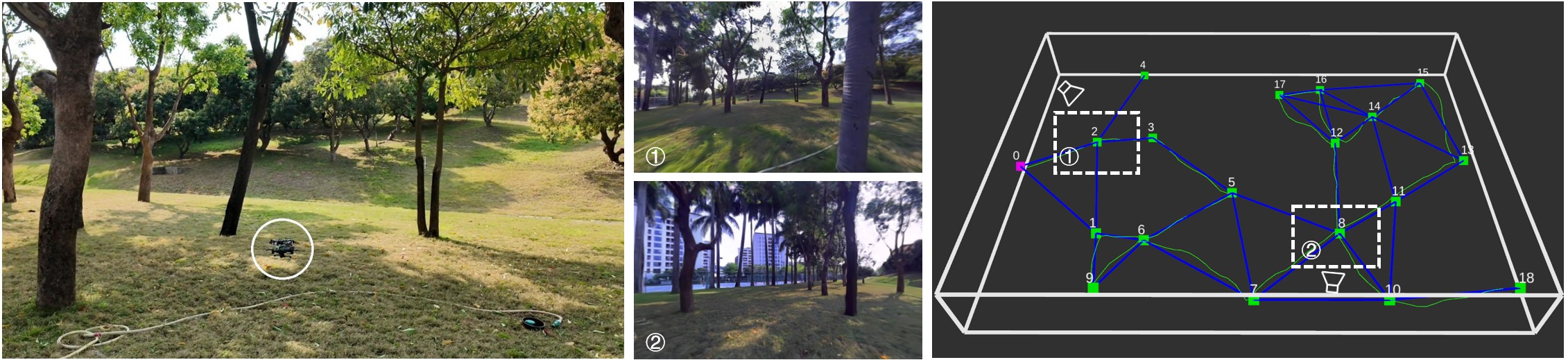}
    \caption{Autonomous exploration experiment in a forest using our palm-sized UAV. Topological map and novel view synthesis results from the trained 3DGS are shown in the right image.}
    \label{fig:exp3}
    \vspace{-1mm}
\end{figure*}

The above-mentioned challenge stems from stringent SWaP constraints that severely limit both sensing and onboard computing capabilities. For sensing, larger UAV platforms can be equipped with high-precision sensors such as LiDAR\cite{EPIC,bubbleExp} and RGB-D cameras\cite{fuel,falcon} to obtain dense and accurate environmental geometric measurements. In contrast, palm-sized micro UAVs can only carry monocular or lightweight multi-camera suites to minimize payload. This leads to inherent defects including scale ambiguity, low depth accuracy, and severe sensing noise. These low-fidelity sensing measurements fail to provide fine-grained environmental geometry information, directly undermining core exploration steps such as reliable map maintenance and frontier boundary extraction.
Beyond sensing limitations, the extremely constrained onboard computing resources further exacerbate these challenges. Traditional exploration frameworks rely heavily on dense volumetric representations to distinguish explored/unknown regions and extract frontiers\cite{fuel}, which incur massive memory and computing overhead. Worse still, low-quality sensing measurements require additional computational resources for noise suppression, creating a fundamental contradiction between soaring computing demand and extremely limited hardware resources.

To address this challenge, we propose a lightweight, resource-efficient autonomous exploration framework, which enables efficient and complete exploration in complex unknown environments using palm-sized UAVs with only lightweight sensing and limited onboard computing resources. We validate our framework on a custom-built palm-sized UAV with a wheelbase of only 11 cm, equipped with a four-fisheye camera setup to achieve omnidirectional environmental perception with minimal payload. To address limitations from coarse depth measurements and constrained computing resources, we propose a frontier node generation method that operates directly on omnidirectional point clouds derived from coarse depth estimation, guiding the UAV toward unexplored regions without explicitly maintaining a dense volumetric representation as in conventional methods. To further ensure complete coverage with minimal overhead, we propose a novel scene representation that augments a sparse topological graph with a depth descriptor for each graph node. This representation compactly encodes explored and unexplored regions and supports dynamic updates with low overhead during exploration, enabling lightweight, efficient, and complete autonomous exploration without the substantial memory and computational overhead imposed by dense map maintenance.

We validate the effectiveness of our proposed framework through extensive real-world flight tests in diverse complex indoor and outdoor scenarios, alongside systematic simulation evaluations.
Experimental results demonstrate that our method achieves efficient and complete exploration of complex environments with coarse sensor measurements and minimal computational overhead.
Furthermore, images collected during exploration enable high-fidelity 3D Gaussian Splatting (3DGS) scene reconstruction, highlighting the practical value of our system.
The main contributions of this paper are summarized as follows:

\begin{itemize}

\item[1)] A lightweight framework that enables complete and efficient exploration of complex unknown environments using a palm-sized micro UAV with low-quality omnidirectional visual input and stringent onboard computing resources.

\item[2)] A frontier node generation method that operates directly on coarse omnidirectional depth, identifying potential unexplored regions without maintaining a volumetric map.

\item[3)] A sparse topological graph representation augmented by a depth descriptor for each graph node, which compactly encodes explored and unexplored regions without dense representation, significantly reducing computational overhead.

\item[4)] Validation of the proposed approach through simulations and real-world experiments, with full system deployment on a custom-built palm-sized quadrotor platform. The code and hardware design will be made open source to benefit the community.

\end{itemize}

\section{Related work}
\label{sec:Related-Work}

\subsection{Scene Representation for Exploration}

Environment representation is the fundamental basis of autonomous unknown environment exploration, which directly determines the efficiency of frontier detection, path planning and the overall exploration pipeline.
Classic frontier-based methods \cite{fuel, bubbleExp, FLARE} adopt occupancy grids for environment representation, in which frontiers are defined as the boundary between known free space and unknown space. Senarathne and Wang\cite{Towards} use OctoMap\cite{OctoMap} as the environment representation, and consider mapped surface voxels neighboring unknown space as surface frontiers, instead of using free-space frontiers to achieve efficient object surface exploration, but it introduces extra computational overhead from surface normal calculation. For some sampling-based exploration methods including NBVP\cite{NBVP} and GBPlanner\cite{GBPlanner}, OctoMap is also used as the core environment representation to perform random sampling in free space. ERRT\cite{ERRT} further adopts UFOMap\cite{ufomap} for scene representation, which extends OctoMap with explicit modeling of unknown, free and occupied three states, and achieves optimization in memory efficiency and query speed. These grid map representations can provide complete geometric information of the environment for frontier detection and path planning, making them the dominant paradigm in current autonomous exploration research. However, these representations inherently suffer from high memory consumption and computational overhead, which severely limits their deployment on resource-constrained aerial robots.

To address the above limitations, a series of lightweight grid-free representation methods have been proposed. FRTree\cite{FRTree} realizes navigation in unknown environments by finding traversable gaps\cite{saferGap} between obstacles, which eliminates the dependence on occupancy grids. It uses a tree structure to store waypoints of free regions.
However, it cannot effectively represent the environment for exploration tasks since it does not maintain the connectivity information of the environment and only supports path search between local tree nodes.
EPIC\cite{EPIC} uses a spatial hash table to store only the observation quality of environmental surfaces, and defines frontiers as poorly-observed voxels adjacent to well-observed voxels. This design abandons the global occupancy grid and improves computational efficiency, but it still requires maintaining a global point cloud map and surface voxels, leading to non-negligible computational resource consumption. Our method replaces the classic dense map with a representation based on sparse topological graph and node depth descriptors, which greatly reduces the computation overhead while ensuring complete coverage of unknown environments.

\begin{figure*}[t]
    \centering
    \includegraphics[width=.99\linewidth]{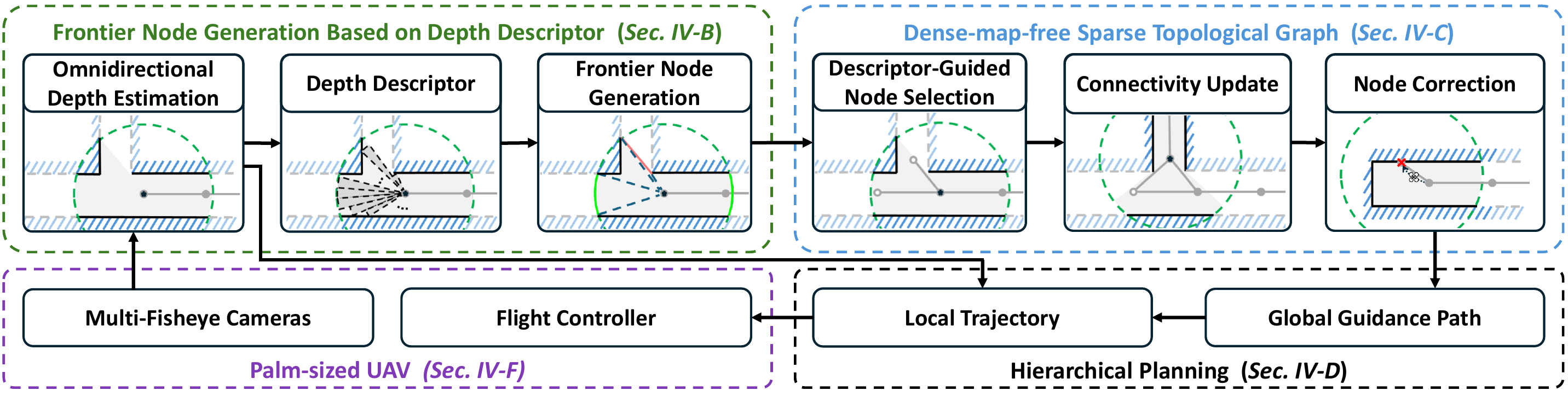}
    \caption{The overview of our proposed exploration approach.}
    \label{fig:framework}
\end{figure*}

\subsection{Exploration with Topological Guidance}

To further improve planning efficiency, some works have introduced topological graph to guide the exploration.
FALCON\cite{falcon} divides the exploration space into multiple sub-regions, updates the connectivity between regions based on occupancy grids, and extracts the environmental topological structure to guide global path planning. EPIC\cite{EPIC} also builds a topological map by decomposing the space into disjoint zones, while it directly updates the connectivity regions using the global point cloud map. Yang et al.\cite{Graph2021} divides the sensor range space into coverage region and frontier region with convex polyhedrons, and builds a topological map by taking these polyhedrons as nodes. Kim et al.\cite{Topo2024} clusters unknown region point clouds to generate Segmented Exploration Regions (SERs) and considers the centroid of SERs as frontiers. It takes the keyframes from LiDAR SLAM as nodes, and constructs a topological graph combined with frontiers to guide the exploration.

Despite the improved planning efficiency, these topology-guided exploration methods follow a two-step "dense mapping first, topological extraction second" paradigm that retains the reliance on maintaining dense maps.
Our method tightly couples topological nodes with frontier detection, representing frontiers as topological nodes to indicate potential unexplored regions.
The proposed scene representation only stores sparse nodes with their types, connectivity and descriptors, eliminating the global dense map maintenance.

\section{System Overview}

The overall framework of the proposed method is illustrated in Fig. \ref{fig:framework}.
During the exploration process, omnidirectional depth is estimated from four fisheye images via a lightweight network, and depth descriptor is calculated on the current coarse point cloud. Frontier nodes are extracted using the depth descriptor of the current frame without explicitly maintaining a dense volumetric representation.
The extracted frontier nodes are used for incremental update of a topological graph, in which only each node's type, connectivity, and depth descriptor are stored. 
Then the hierarchical planning is performed on the constructed topological graph to guide the exploration. The entire exploration process is terminated when no frontier nodes remain in the topological graph.

\section{Proposed Method}
\label{sec:Proposed Method}

\subsection{Scene Representation}

To avoid the maintenance of global occupancy grid maps and dense point cloud maps, the global environment is represented solely by a sparse graph, where each node stores a local depth descriptor that captures the geometric characteristics of its surrounding region, providing enough information for scene representation and new frontier node generation. 
As defined in \cite{frontier}, a frontier is the boundary between known and unknown regions. Instead of extracting explicit geometric boundaries, our method directly generates new frontier nodes in the potential boundary regions using depth descriptors of the current node and historical nodes, guiding the UAV toward unexplored regions.

The global environment is represented by an undirected sparse topological graph, which is formally defined as: $ \mathcal{G} = \left( \mathcal{N}, \mathcal{E} \right) $, 
where $\mathcal{N}$ denotes the set of nodes, and $\mathcal{E}$ denotes the set of undirected edges in the graph.
The node set $\mathcal{N}$ consists of two types: \textit{WAYPOINT}-type nodes $\mathcal{N}^w$ that correspond to visited nodes (referred to as waypoint nodes), representing local traversable space, and \textit{FRONTIER}-type nodes $\mathcal{N}^f$ that are unvisited nodes (referred to as frontier nodes), indicating potential unexplored regions. Each node $N_i \in \mathcal{N}$ stores necessary information for exploration:
\begin{itemize}
\item[1.] $\textit{node\_type}$: The type identifier of the node, with the value of either $\textit{WAYPOINT}$ or $\textit{FRONTIER}$;
\item[2.] $\boldsymbol{P}_i \in \mathbb{R}^3$: The position coordinate of the node in the world coordinate system;
\item[3.] $\mathcal{D}_i$: The depth descriptor of the node. The range space of the current frame point cloud is partitioned into multiple uniform fan-shaped regions, and each entry in the depth descriptor stores the minimum depth value of valid points within the corresponding region;
\item[4.] $\mathcal{N}^f_{adj,i}$: The set of all \textit{FRONTIER}-type nodes directly connected to the current node;
\item[5.] $\mathcal{N}_{adj,i}$: The set of all adjacent nodes directly connected to the current node;
\item[6.] $\mathcal{C}_{adj,i}$: The set of traversal costs corresponding to the adjacent nodes. The connectivity between nodes is updated based on local line-of-sight visibility checks, ensuring that there is no obstacle occlusion between adjacent nodes. Thus for computational simplification, the traversal cost $C_{ij}$ from node $N_i$ to its adjacent node $N_j$ is defined as the Euclidean distance between the two nodes:
$ C_{ij} = \left\| \boldsymbol{P}_i - \boldsymbol{P}_j \right\|_2 $.
\end{itemize}

Each edge $e_{ij} \in \mathcal{E}$ corresponds to a pair of connected nodes $N_i$ and $N_j$, with its weight set as the traversal cost $C_{ij}$ defined above. The topological graph is maintained via an adjacency list structure. 

When the UAV moves close to a target frontier node, the type of that node is immediately converted from FRONTIER to WAYPOINT. Then new frontier node generation and connectivity update procedures are executed simultaneously. New frontier nodes are generated based on the depth descriptor of the current newly-converted waypoint node, and the connectivity of the current node in the graph is also updated to complete the incremental update.

\subsection{Frontier Node Generation Based on Depth Descriptor}

\begin{figure}[t]
    \centering
    \includegraphics[width=.95\linewidth]{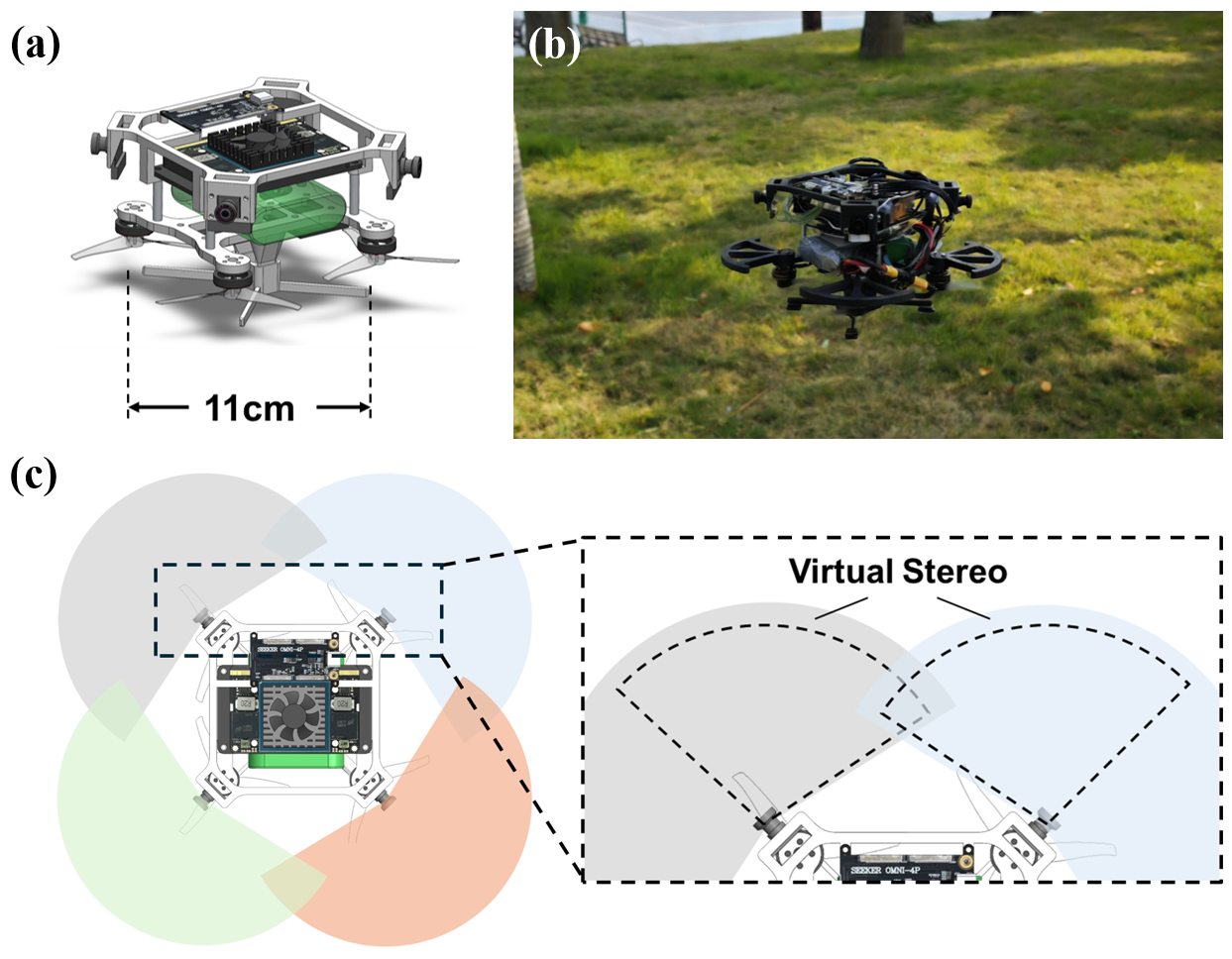}
    \caption{(a) The mechanical model of the palm-sized UAV. (b) The UAV equipped with propeller guards, performing autonomous flight in the forest scenario. (c) Four-fisheye cameras and the virtual stereo for depth estimation.}
    \label{fig:fisheyestereo}
    % \vspace{-0.3cm}
\end{figure}

\begin{figure*}[t]
    \centering
    \includegraphics[width=.95\linewidth]{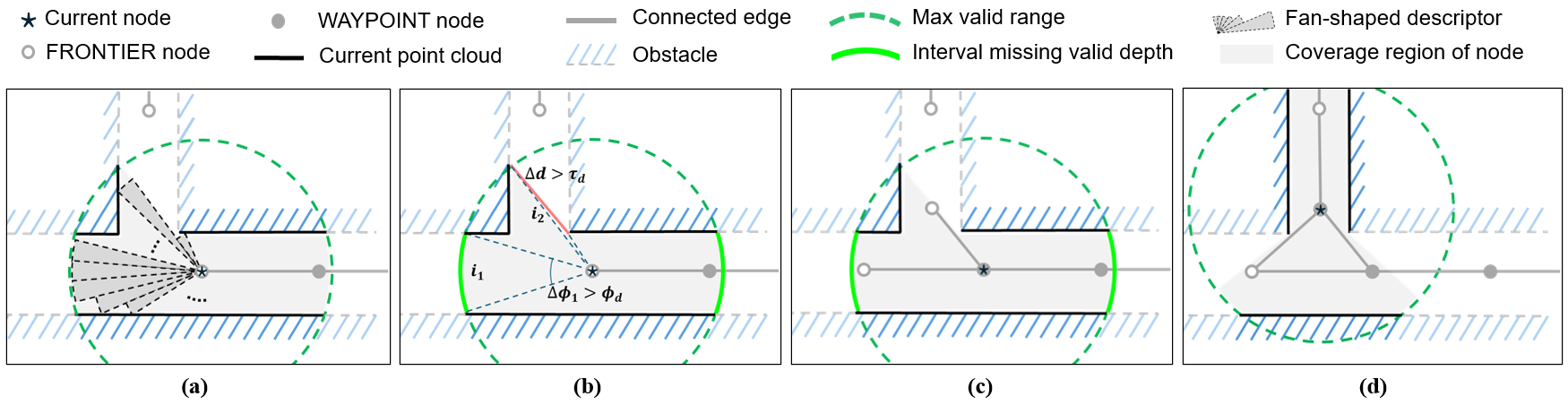}
    \vspace{-0.1cm}
    \caption{Frontier node generation and topological map construction. In (a) and (b), two traversable intervals are calculated, and frontier nodes are generated in (c). The right interval containing the last waypoint node is not used for new frontier node generation. Once moving to the next target node, the connectivity between nearby nodes is updated in (d).}
    \label{fig:topo}
    \vspace{-0.2cm}
\end{figure*}

\subsubsection{Omnidirectional Depth Estimation}

For frontier node generation and local obstacle avoidance, it is essential to estimate depth information from the onboard visual input. Inspired by \cite{omninxt}, a four-fisheye camera configuration is adopted to achieve omnidirectional depth perception. Despite the similar sensor setup, our UAV platform features a more lightweight and compact design compared with \cite{omninxt}, with detailed descriptions in Sec.\ref{sec:platform}.
As shown in Fig. \ref{fig:fisheyestereo} (c), the four fisheye images are undistorted and split into multiple virtual stereo pairs. For each virtual stereo pair, a stereo matching network\cite{hitnet} is adopted for stereo depth estimation. The depth maps from virtual stereo pairs then are fused into full omnidirectional point cloud via pre-calibrated camera extrinsic parameters.

It is noted that although this omnidirectional depth estimation pipeline can achieve over 10 FPS on the NVIDIA Orin NX onboard platform\cite{omninxt}, the depth estimation frame rate is set to only 2 FPS in real-world experiments, by which the onboard computing resource consumption is further reduced while the requirements of autonomous exploration can be satisfied.

\subsubsection{Depth Descriptor}

As mentioned before, frontier nodes are generated on current frame point cloud.
To address the limited accuracy of depth estimation, the current coarse point cloud is first represented as a depth descriptor, and frontier nodes are then extracted directly from this descriptor to identify potential unexplored regions around the current position rather than computing explicit boundary.

Let $\mathcal{W}$ denote the world frame and $\mathcal{B}$ denote the UAV body frame. For computational convenience, a virtual coordinate frame $\mathcal{C}$ is established, whose origin coincides with the origin of $\mathcal{B}$, and axes are parallel to those of $\mathcal{W}$. The current frame point cloud is first transformed from the UAV body frame $\mathcal{B}$ to the virtual frame $\mathcal{C}$. Let $d_{max}$ denote the maximum valid sensing distance of the depth estimation, and $h$ denote the height range for valid point extraction. The valid point cloud $\mathcal{P}_{v}$ of the current frame is extracted as:
\begin{equation}
\mathcal{P}_{v} = \left\{ \boldsymbol{p} \in {^\mathcal{C}}\mathcal{P} \mid -\frac{h}{2} < p_z < \frac{h}{2} \text{ and } \|\boldsymbol{p}\|_2 < d_{max} \right\}
\end{equation}
where ${^\mathcal{C}}\mathcal{P}$ is the point cloud in the virtual frame $\mathcal{C}$, and $p_z$ is the z-coordinate of the point $\boldsymbol{p}$.

Subsequently, the point cloud range space around the virtual frame origin is divided into $n$ uniform fan-shaped regions shown in Fig. \ref{fig:topo}(a), with the angular resolution set as $\theta$ (in degrees), where $n = 360^{\circ}/\theta$. For each fan-shaped region $j$ ($j = 1, 2, ..., n$), the minimum depth value $d_j$ of valid points within the region is computed and stored in the depth descriptor $\mathcal{D}_i = \left[ d_1, d_2, ..., d_n \right]$ of the current node. If no valid point exists in the $j$-th region, the corresponding depth value is set to $d_{max}$.

\subsubsection{Frontier Node Generation}

Candidate frontier nodes are identified based on two types of characteristics in the depth descriptor $\mathcal{D}_i$: (1) angular ranges with missing valid depth information, i.e., consecutive regions with $d_j = d_{max}$, and (2) angular ranges with significant depth discontinuities between near regions.
Let $\phi_d$ denote the angular range threshold for missing valid depth, and $\tau_d$ denote the depth discontinuity threshold. 
As shown in Fig. \ref{fig:topo} (b), a set of traversable angular intervals $\mathcal{I} = \left\{ \boldsymbol{i}_k \right\}$ is obtained.
Each interval in the set is represented by a 2D vector $\quad \boldsymbol{i}_k = \left[ n_{k,left}, n_{k,right} \right]$ containing the indices of the interval boundaries, where $n_{k,left}$ and $n_{k,right}$ are the indices of the left and right boundaries of the $k$-th interval.
If an extracted angular interval with missing valid depth is larger than a threshold, it will be equally split into sub-intervals to replace the original range for frontier node calculation.
To avoid generating frontier nodes within the last waypoint node's coverage, the heading angle of the previous waypoint node in the current virtual frame is calculated. Any angular interval containing this heading angle is removed from $\mathcal{I}$ for preliminary filtering of frontier node generation.

The position of the candidate frontier node $\boldsymbol{w}_k$ corresponding to interval $\boldsymbol{i}_k$ is computed as:

\begin{equation}
\mathbf{p}_{j} =
\begin{bmatrix}
d_{n_{k,j}}\cos\left(n_{k,j} \cdot \theta\right)\\
d_{n_{k,j}}\sin\left(n_{k,j} \cdot \theta\right)\\
z_p

\end{bmatrix}, j = left, right
\end{equation}

\begin{equation}
\boldsymbol{w}_{k} = (\mathbf{p}_{left} + \mathbf{p}_{right})/2
\end{equation}
where $\mathbf{p}_j$ denotes the point computed using the minimum depth corresponding to the interval left or right boundary, and $z_p$ is the z-coordinate of the current node. The candidate node is considered as the midpoint of these two boundary points.

This candidate frontier node generation approach detects potential unexplored regions from the current-frame point cloud and descriptors. It requires no detailed geometric input, avoids explicit frontier extraction, and is suited for omnidirectional point clouds with limited depth estimation accuracy.

\subsection{Dense-map-free Sparse Topological Graph}

\subsubsection{Descriptor-Guided Node Selection}

Since no global dense map is maintained and candidate frontier nodes are generated solely based on the current frame point cloud, the generated nodes may lie within previously visited regions. Although preliminary filtering using the last waypoint node is implemented during frontier node generation, this constraint is still insufficient for redundancy suppression.

For each generated candidate frontier node $\boldsymbol{w}_k$, all historical nodes $\boldsymbol{N}_i \in \mathcal{N}$ whose Euclidean distance to $\boldsymbol{w}_k$ is less than the valid sensing distance $d_{max}$ are retrieved from the topological graph $\mathcal{G}$. 
The heading angle and distance from each historical node $\boldsymbol{N}_i$ to $\boldsymbol{w}_k$ are then computed, and the heading angle is mapped to the corresponding index in depth descriptor $\mathcal{D}_i$ stored in $\boldsymbol{N}_i$.
Subsequently, the depth values within an angular window $\Delta\theta$ centered at the target heading angle are checked. If all depth values within the window are greater than the distance between $\boldsymbol{w}_k$ and $\boldsymbol{N}_i$, the candidate node is determined to be within the already observed range of $\boldsymbol{N}_i$ and is removed from the candidate frontier node set.

The remaining candidate nodes after the above filtering are added to the topological graph as frontier nodes, and the connectivity and edge costs between the current node and these frontier nodes are updated simultaneously. This filtering process computes visibility check from historical node descriptors, further suppressing redundant frontier nodes in explored regions.

\subsubsection{Connectivity Update}

Once the UAV visits a frontier node and the type of the node is changed to WAYPOINT, the connectivity of the current waypoint node in the graph will be updated.
% The connectivity of the current node in the graph will be updated once the UAV visited .
The connectivity between nodes are also updated based on the depth descriptor of the current node via local line-of-sight visibility. The nodes whose Euclidean distance to the current node is less than $d_{max}$ are retained. The nodes are transformed from the world inertial frame $\mathcal{W}$ to the current virtual frame $\mathcal{C}$, and corresponding heading angle $\theta_p$ and distance $r_p$ are calculated. The depth descriptor of the current node is queried within an angular window $\Delta\theta$ centered at $\theta_p$. If all depth values within the window are greater than $r_p$, unobstructed line-of-sight connectivity between the two nodes is determined, and a corresponding edge is added in the topological graph.

\subsubsection{Node Correction}

\begin{figure}[t]
    \centering
    \includegraphics[width=.98\linewidth]{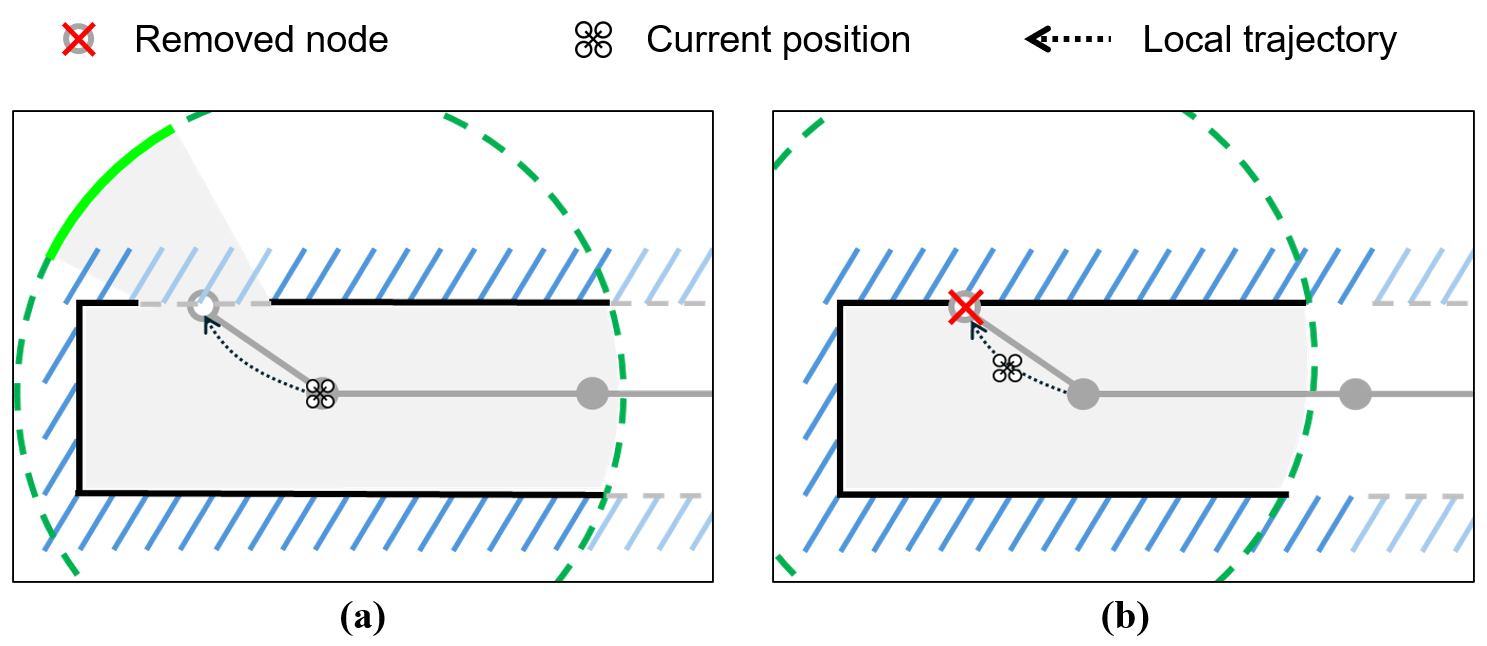}
    \caption{The node correction for erroneously generated node. (a) A frontier node generated within obstacle due to inaccurate depth estimation. (b) The removal of the node.}
    \label{fig:correction}
    \vspace{-0.3cm}
\end{figure}

The limited accuracy of omnidirectional depth estimation, especially in challenging scenarios such as textureless regions and reflective surfaces, may introduce erroneous depth measurements, which further leads to the generation of invalid frontier nodes. 
As shown in Fig. \ref{fig:correction} (a), a frontier node may be incorrectly generated inside an obstacle wall due to erroneous depth estimation. To address this issue, a node correction mechanism is introduced.

The core design leverages the inherent characteristic of stereo depth estimation: its relative accuracy improves significantly as the observation distance decreases. When the UAV navigates toward a target frontier node $N_k^f$, real-time validity check of the target node is performed using the depth from the current frame.
The validity check procedure follows a consistent logic with the depth coverage check described in the previous section. First, the depth descriptor vector $\mathcal{D}_{curr}$ is computed from the current frame point cloud. If the target node is determined to be located outside the valid observable region characterized by $\mathcal{D}_{curr}$, the node will be identified as invalid. This invalid node is immediately removed from the topological graph, and a new global path is re-planned to sustain continuous autonomous exploration.

\subsection{Hierarchical Exploration Planning}

\subsubsection{Global Path Guidance on Topological Graph}

Starting from the current waypoint node, the shortest path that traverses all frontier nodes in the topological graph is generated. The cost between any two nodes is calculated as the shortest path distance on the graph, solved via a graph-based A* algorithm. Inspired by previous works \cite{fuel,EPIC}, the traversal problem is formulated as an Asymmetric Traveling Salesman Problem (ATSP) \cite{ATSP}.
To simplify the computation and reduce computational overhead, if the current node has directly connected frontier node neighbors, the nearest frontier node will be set as the next target for efficient local exploration advance. 
When the current node has no directly connected frontier nodes, the ATSP is solved to generate a global path to guide the exploration.

\subsubsection{Local Trajectory Planning}

Once the global guidance path composed of a sequence of topological nodes from the constructed topological graph is obtained, dynamically feasible and collision-free trajectories are generated for the UAV in real time. A local occupancy grid map is constructed only using the real-time point cloud of the current frame. We adopt EGO-Planner\cite{egoplanner} for local trajectory optimization.
The local grid map operates within a limited sliding window without persistent global storage to reduce computational and memory overhead.
Since the full omnidirectional FoV of the depth estimation eliminates the need for heading adjustment, the heading angle is thus fixed to 0 throughout the trajectory optimization process.

\subsection{Scene Reconstruction}
Since the proposed exploration framework does not maintain a global dense map, to qualitatively validate the coverage of the explored environment in real-world experiments, raw fisheye images captured during the autonomous exploration process are stored for offline scene reconstruction.
To accommodate the large-FoV fisheye image input, the Fisheye-GS method\cite{FisheyeGS} is adopted for scene reconstruction.
In order to reduce the influence of inconsistent exposure of cameras in different poses and eliminate floaters, we use 3D bilateral grids \cite{bilarf} to disentangle inconsistent image signal processing (ISP) during 3DGS training. The final reconstructed 3D scene is used to visualize and verify the convergence of the environment.

\subsection{Palm-sized UAV Platform}
\label{sec:platform}

To validate the performance of the proposed method in real-world unknown environment exploration, we built a palm-sized, omnidirectional perception quadrotor UAV platform shown in Fig. \ref{fig:fisheyestereo}.
A four-fisheye camera configuration similar to \cite{omninxt} is adopted for full-environment perception, where each fisheye camera has an approximate 200$^{\circ}$ field of view (FoV) to provide omnidirectional perception. Benefiting from our compact and lightweight structural design, the UAV achieves a wheelbase of only 11 cm and a total weight of around 400 g, which is smaller and lighter than the platform presented in \cite{omninxt}.
For onboard computing and flight control, the UAV is equipped with an NVIDIA Orin NX 16 GB and a NxtPx4 autopilot. The full pipeline of the proposed framework, including depth estimation and exploration path planning, is executed entirely in real time on board.
For state estimation, a modified visual-inertial odometry (VIO) framework based on \cite{d2slam} is adopted, which fuses measurements from the four fisheye cameras and an onboard IMU to achieve state estimation. To strictly constrain the power consumption of the onboard computer, we do not use the MAXN full-power mode of the Orin NX to unlock the full computing performance as done in \cite{omninxt}. Instead, the 25 W power mode is applied, with the upper limit of the GPU frequency moderately increased on this basis, while the final operating frequency is still significantly lower than the nominal GPU frequency under the MAXN mode. This configuration is adopted to verify the real-time performance and feasibility of the proposed method under constrained onboard computing resources.

\section{Experiments and evaluation}
\label{sec:Results}

\subsection{Simulation Evaluations}
\begin{figure}[t]
    \centering
    \begin{subfigure}[t]{.53\linewidth}
        \centering
        \includegraphics[width=1\textwidth]{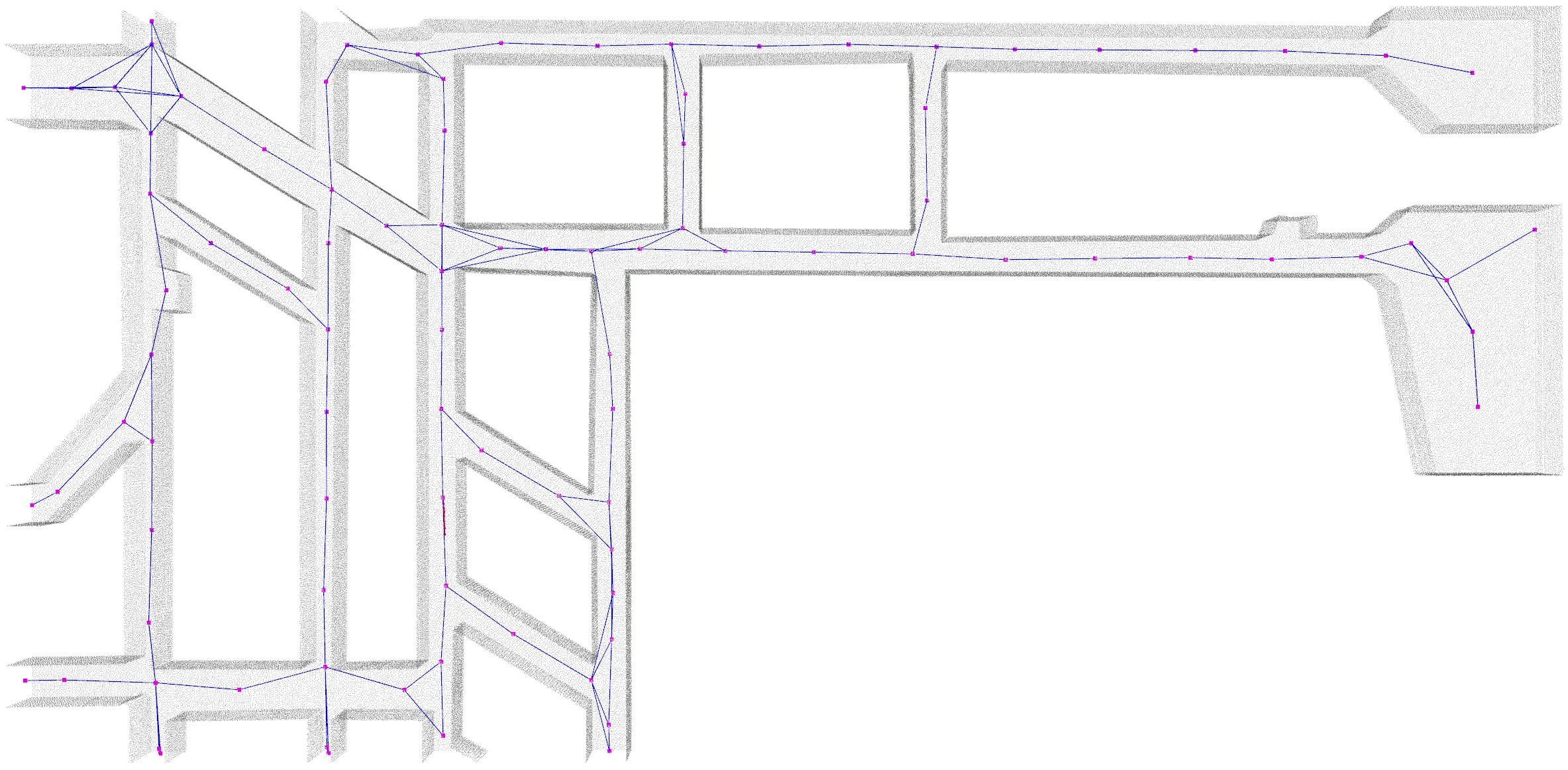}
        \caption{Tunnel}
    \end{subfigure}
    \begin{subfigure}[t]{.45\linewidth}
        \centering
        \includegraphics[width=1\textwidth]{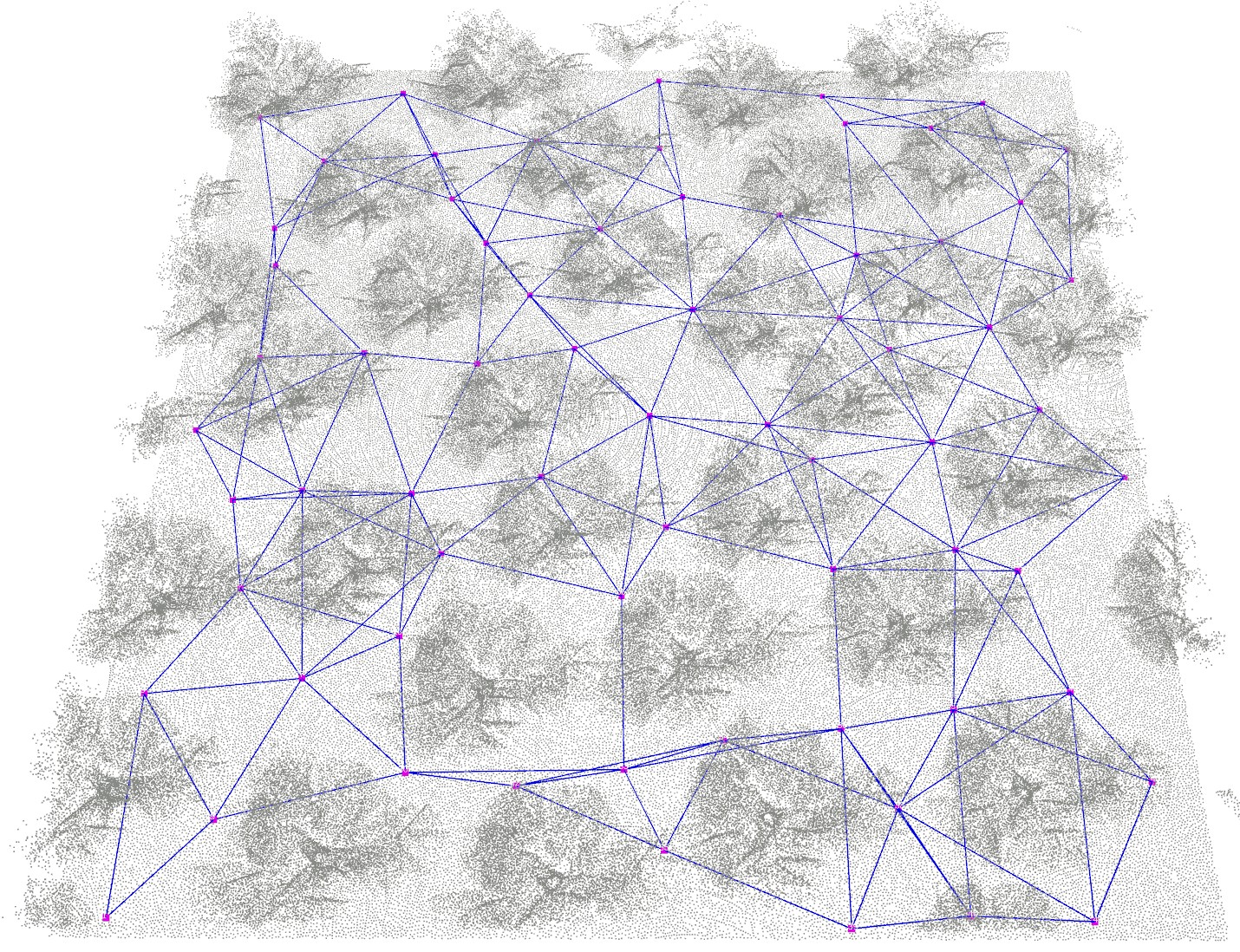}
        \caption{Forest}
    \end{subfigure}
    \caption{The constructed topological graph of our method in two simulation scenes.}
    % \vspace{-0.3cm}
    \label{fig:simu}
\end{figure}

\begin{figure}[t]
    \centering
    \begin{subfigure}[t]{.75\linewidth}
        \centering
        \includegraphics[width=1\textwidth]{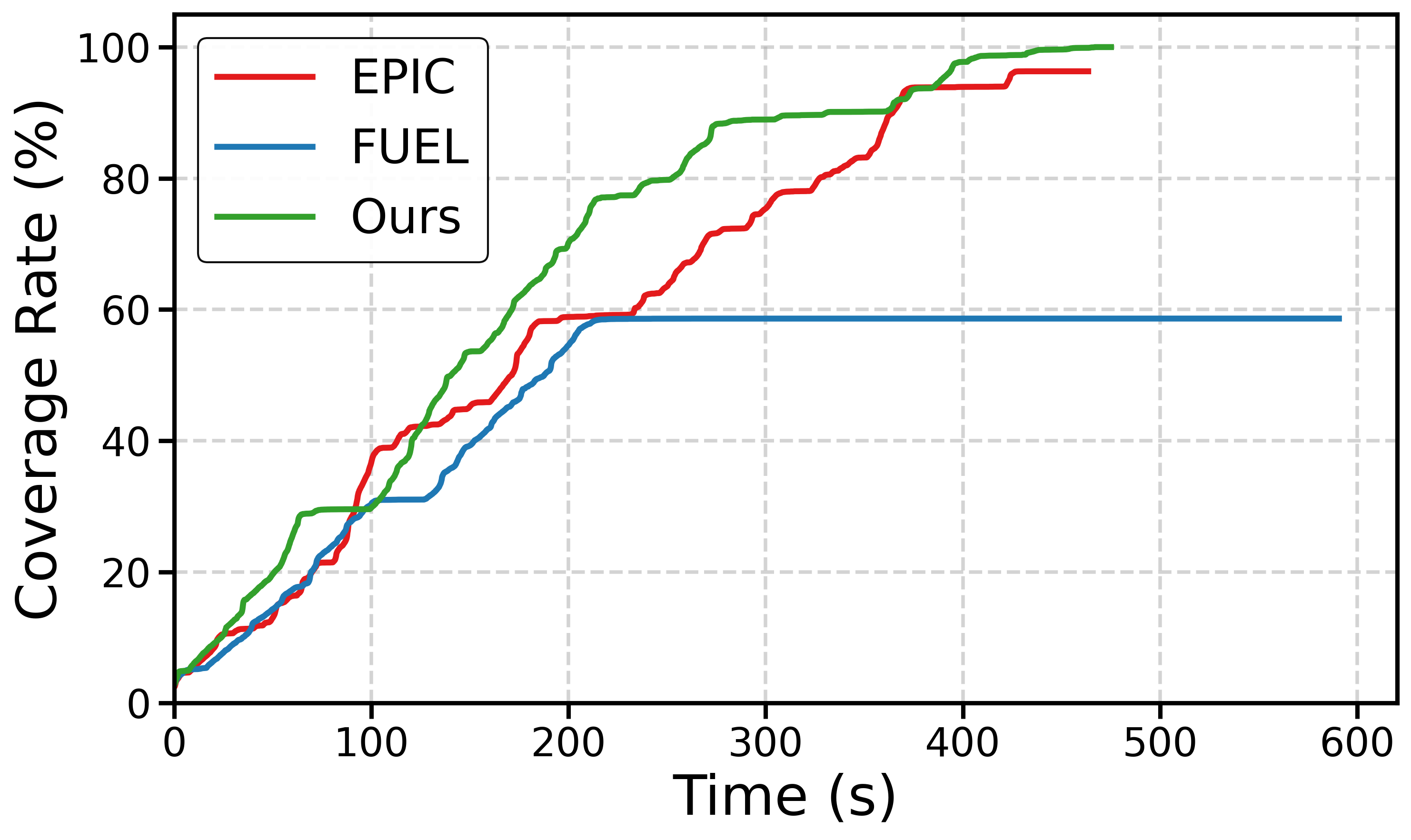}
        \label{fig:matrics_tunnel}
    \end{subfigure}
    \vspace{-0.28cm}
    \begin{subfigure}[t]{.75\linewidth}
        \centering
        \includegraphics[width=1\textwidth]{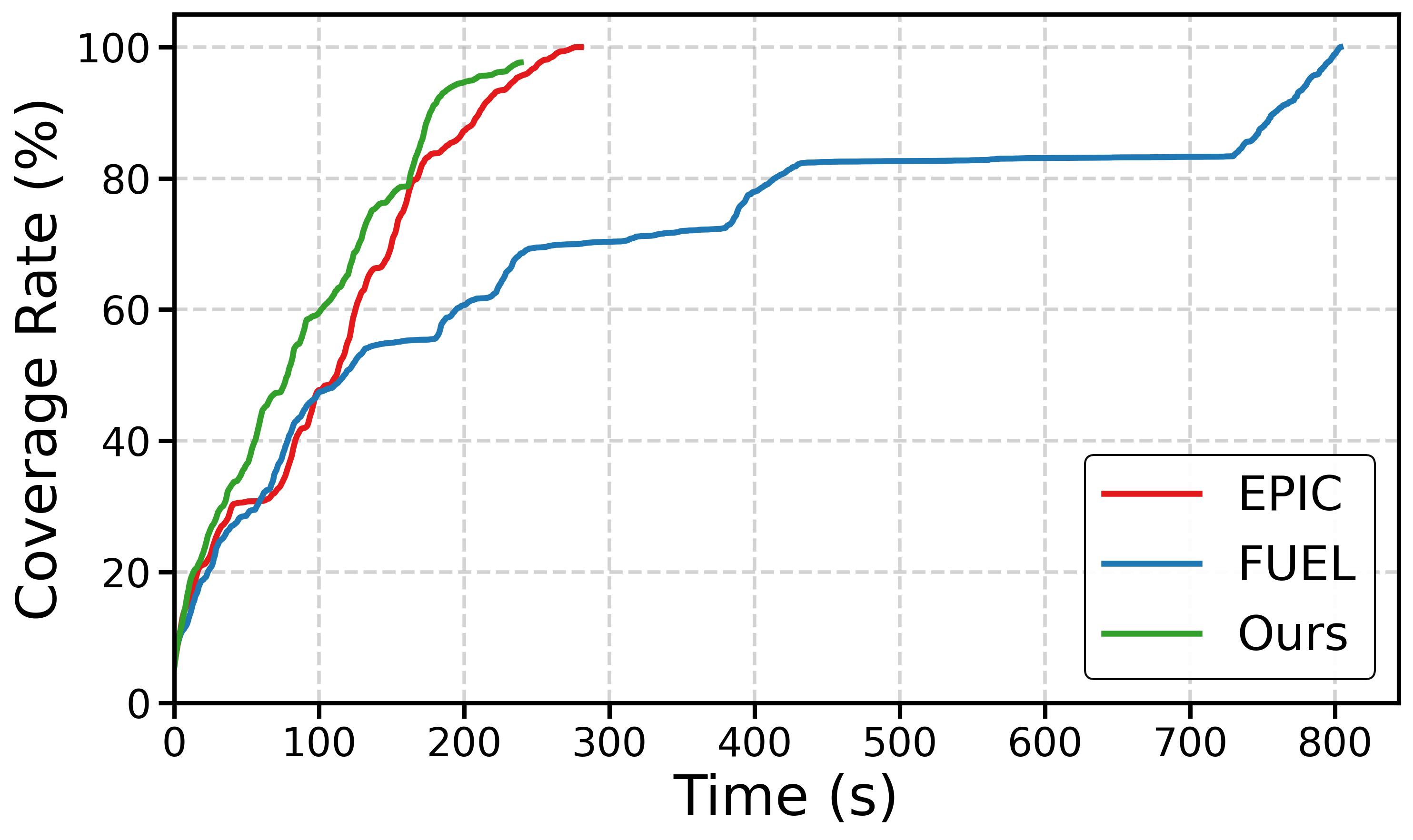}
        \label{fig:matrics_tunnel}
    \end{subfigure}
    \vspace{-0.28cm}
    \caption{The exploration progress of the three methods in tunnel (upper) and forest (lower) scenarios.}
    \label{fig:metrics}
\end{figure}

Simulation experiments are conducted on the MARSIM simulator\cite{marsim}, with two simulated environments provided by \cite{cmusim} for autonomous exploration: a tunnel scene 120 m $\times$ 53 m $\times$ 3 m and a forest scene 50 m $\times$ 50 m $\times$ 2 m shown in Fig. \ref{fig:simu}. The state-of-the-art autonomous exploration framework FUEL \cite{fuel} and EPIC\cite{EPIC} are selected as the baselines for comprehensive performance comparison.
To simulate the inherent noise of omnidirectional depth estimation in real-world deployment, additional noise is added to the point cloud generated by the simulation sensor throughout the exploration process. All methods are run on an Intel Core i7-14700KF CPU with 32 GB RAM in Ubuntu 20.04.
Each method is run three times in each scenario, with the maximum velocity set to 2 m/s and the maximum acceleration set to 2 m/s$^2$.
The exploration progress comparison between these methods is visualized in Fig. \ref{fig:metrics}.

\subsubsection{Exploration Efficiency}
In both tunnel and forest scenes, our method achieves comparable exploration efficiency with EPIC.
In the tunnel scenario, EPIC missed full exploration of narrow tunnel segments by neglecting small frontier voxel clustering, while FUEL failed to finish exploration within the time limit due to frequent back-and-forth movements. In the forest scene, our method shows minor incomplete coverage in some small areas between the exploration boundary and the trees close to it, caused by the simplified scene representation that omits fine-grained details, but still retains overall high coverage rate and exploration efficiency. FUEL completed the exploration in this scenario but with low efficiency due to its frequent back-and-forth movements and high computation time of global path planning when the explored space is large.

\subsubsection{Memory Consumption}
The comparison of memory consumption of scene representations is shown in Table \ref{table:memory}. For scene representation, we evaluate FUEL's global uniform grid map, EPIC's observation map with global point cloud map, and our method's topological graph with sliding-window local grid map.
The global uniform occupancy grid map maintained by FUEL incurs a remarkably high memory usage. Although EPIC's observation map is lightweight, its simultaneously maintained global point cloud map still occupies non-negligible memory. In comparison, our topological graph with a sliding-window grid map achieves extremely low memory consumption.

\subsubsection{Computation Time}
We also compare the average computation time per planning iteration across different methods shown in Table \ref{table:time}. Each planning iteration is decomposed into three sequential functional modules: map update, global path planning, and local trajectory generation, where the map update module includes both scene representation update and frontier computation.
The results show that our method exhibits significant advantages in both memory consumption and computation time.

\begin{table}[t]
\centering
\caption{Memory Consumption of Map (MB)}
\label{table:memory}
\begin{tabular}{lccc}
\hline
\vspace{-0.28cm}
\\ \hline
\textbf{Scene}  & FUEL\cite{fuel}   & EPIC\cite{EPIC}          & ours               \\ \hline
Tunnel & 594.41 & 0.52 + 123.85 & \textbf{0.16 + 0.71} \\
Forest & 233.65 & 1.50 + 248.22   & \textbf{0.13 + 0.71} \\ \hline
\vspace{-0.28cm}
\\ \hline
\end{tabular}
\end{table}

\begin{table}[!h]
\setlength{\tabcolsep}{3pt}
\centering
\caption{Computation Time per Planning Iteration (ms)}
\label{fig:time}
\label{table:time}
\begin{tabular}{llcccc}
\hline
\vspace{-0.28cm}
\\ \hline
% \toprule
\textbf{Scene}  & \textbf{Method} & \textbf{Map Update} & \textbf{Global Path} & \textbf{Local Traj.} & \textbf{Total}           \\ \hline
       & FUEL\cite{fuel}   & 17.48     & 457.65     & 0.65            & 475.78         \\
Tunnel & EPIC\cite{EPIC}   & 35.50    & 20.30     & 10.79           & 66.59         \\
       & ours   & 6.78     & 0.03       & 0.11            & \textbf{6.92} \\ \hline
       & FUEL\cite{fuel}   & 39.53      & 426.31     & 1.37            & 467.21          \\
Forest & EPIC\cite{EPIC}   & 51.75    & 18.05     & 20.84           & 90.64          \\
       & ours   & 4.84     & 0.02        & 0.16            & \textbf{5.01} \\ \hline
\vspace{-0.28cm}
\\ \hline
\end{tabular}
\end{table}

\subsection{Real-World Experiments}

\begin{figure}[t]
    \centering
    \includegraphics[width=.99\linewidth]{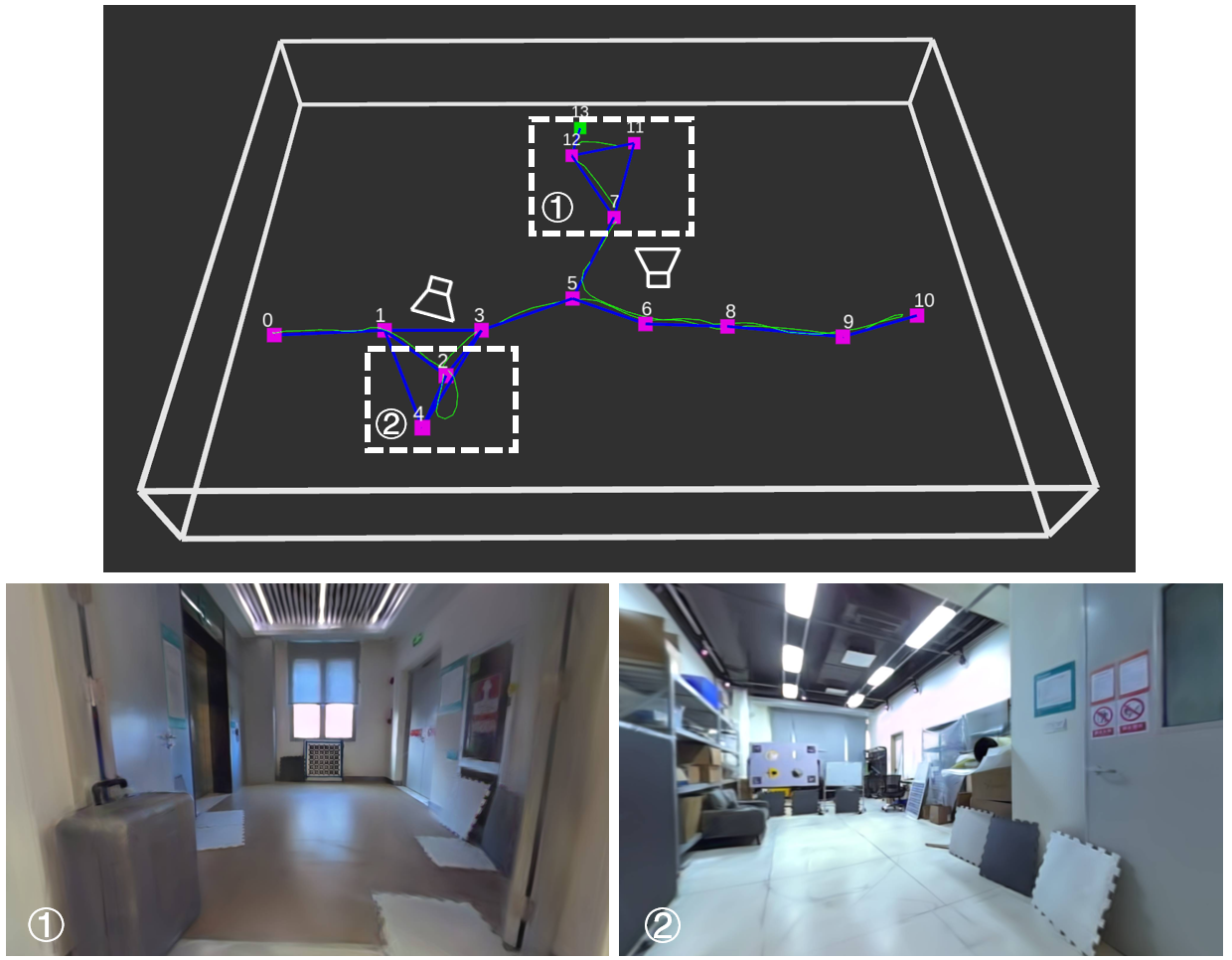}
    \caption{Experiment in an indoor corridor. Topological map and novel view synthesis rendered from the trained 3DGS are shown in the figure.}
    \label{fig:exp1}
\end{figure}

\begin{figure}[t]
    \centering
    \includegraphics[width=.99\linewidth]{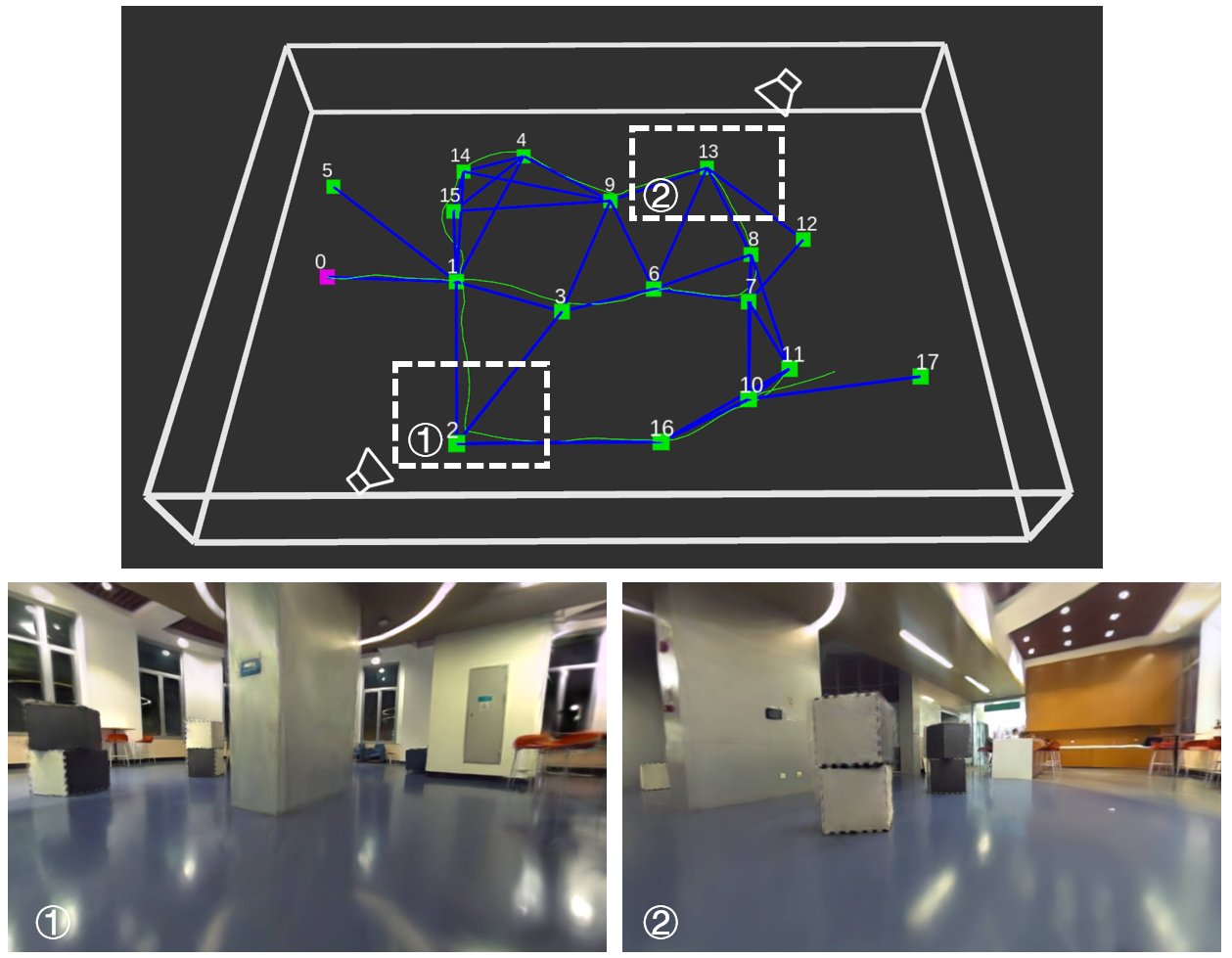}
    \caption{Real-world experiment in an unstructured break room. Final topological map and novel view synthesis rendered from the trained 3DGS are shown in the figure.}
    \label{fig:exp2}
    \vspace{-5mm}
\end{figure}

Autonomous exploration experiments are conducted in three different scenes. The maximum velocity is set to 0.6 m/s and the maximum acceleration is set to 0.5 m/s$^{2}$ among all scenes. The first scenario is an indoor corridor, with the topological map and reconstruction results shown in Fig. \ref{fig:exp1}. The test environment is an enclosed space bounded within a box of 16 m $\times$ 12 m $\times$ 2 m, which consists of a long corridor and two independent rooms. The exploration process took 71.5 s with a 29.8 m trajectory.
During the autonomous exploration process, the UAV incrementally built a topological graph that achieves coverage of the entire test environment. The 13th node was mistakenly generated in an obstacle due to incorrect depth estimation and was removed through the node correction mechanism. The rendering results of the 3DGS reconstruction are also presented in the figure, demonstrating the completeness of exploration.

To verify the feasibility of the proposed method in unstructured environments, the second scenario is a cluttered break room with dimensions of 14 m $\times$ 10 m $\times$ 2 m shown in Fig. \ref{fig:exp2}. The UAV accomplished the exploration task in 80.1 s with a 32.5 m trajectory.
We also conducted exploration tests in a forest with a  15 m $\times$ 11 m $\times$ 2 m exploration space shown in Fig. \ref{fig:exp3} to validate our method in natural environments. The exploration process took 135.0 s with a 54.1 m trajectory. Detailed experimental results and visualizations are available in the supplementary video.

\section{Conclusion}

In this paper, we present a lightweight omnidirectional vision-based autonomous exploration framework for SWaP-constrained UAVs.
Our framework represents potential unexplored areas via frontier nodes with depth descriptors, and models the explored environment through an incrementally constructed sparse topological map with dynamically updated node connectivity, eliminating the need to maintain global dense representations. 
Extensive simulation and real-world experiments validate the feasibility of the proposed framework. We realize fully onboard real-time autonomous exploration on a palm-sized UAV in real-world scenarios.

While the proposed framework achieves efficient and lightweight autonomous exploration, the depth descriptor and topological graph are designed for 2.5D exploration scenarios. For future work, we will generalize the pipeline to full 3D environment exploration by extending the fan-shaped depth descriptor to 3D representation and updating the topological graph to support 3D node connectivity maintenance.

% \section*{ACKNOWLEDGMENT}

% %%%%%%%%%%%%%%%%%%%%%%%%%%%%%%%%%%%%%%%%%%%%%%%%%%%%%%%%%%%%%%%%%%%%%%%%%%%%%%%%

% References are important to the reader; therefore, each citation must be complete and correct. If at all possible, references should be commonly available publications.

% \balance
% \IEEEtriggeratref{8}
\bibliographystyle{IEEEtran}
\bibliography{ref.bib}

\end{document}